\documentclass[11pt]{article}

\usepackage[letterpaper,margin=1in]{geometry}
\usepackage[T1]{fontenc}
\usepackage{lmodern}
\usepackage{microtype}
\usepackage{amsmath,amssymb,amsthm,mathtools}
\usepackage{graphicx}
\usepackage{booktabs}
\usepackage{array}
\usepackage[small,labelfont=bf]{caption}
\usepackage{xcolor}
\usepackage[round,authoryear]{natbib}
\usepackage[colorlinks=true,linkcolor=blue!50!black,citecolor=blue!50!black,urlcolor=blue!50!black]{hyperref}

\newcommand{\memStepFiftyNine}{400}

\newcommand{\meanDlogSFiftyNine}{1{,}960}

\newcommand{\posNatMFiftyNine}{9}

\newcommand{\posNatSFiftyNine}{1}

\newcommand{\bothSFiftyNine}{20}

\newcommand{\plateauFiftyNine}{18}

\newcommand{\firedSFiftyNine}{20}

\newcommand{\nullDlogMaxFiftyNine}{0.39}

\newcommand{\memStepNinetySeven}{200}

\newcommand{\meanDlogSNinetySeven}{1{,}237}

\newcommand{\posNatMNinetySeven}{3}

\newcommand{\posPermMNinetySeven}{0}

\newcommand{\bothSNinetySeven}{19}

\newcommand{\plateauNinetySeven}{17}

\newcommand{\firedSNinetySeven}{19}
\newcommand{\nullDlogNinetySeven}{0.24}
\newcommand{\nullNatNinetySeven}{0.05}

\newcommand{\rhoMultNinetySeven}{-0.08}

\newcommand{\grokMinOneThirteen}{2{,}500}
\newcommand{\grokMaxOneThirteen}{3{,}300}

\newcommand{\posDlogMOneThirteen}{15}

\newcommand{\meanDlogSOneThirteen}{405}

\newcommand{\bothSOneThirteen}{20}

\newcommand{\plateauOneThirteen}{12}
\newcommand{\afterGrokOneThirteen}{7}

\newcommand{\multNets}{60}
\newcommand{\bothMTotal}{60}
\newcommand{\bothSTotal}{59}
\newcommand{\plateauSmallTotal}{35}
\newcommand{\firedSmallTotal}{39}

\newcommand{\beforeMemSFive}{10}

\newcommand{\nullUntrainedSFive}{0.40}

\newcommand{\nullGaussSFive}{0.08}
\newcommand{\memStepSFive}{900}
\newcommand{\rhoSFive}{+0.53}
\newcommand{\pRhoSFive}{0.02}
\newcommand{\nGrokDSixteen}{19}

\newcommand{\posZpPermZFiveZFive}{1}
\newcommand{\firedZpPermZFiveZFive}{20}

\newcommand{\posOrigZEightZNine}{6}

\newcommand{\posMZEightZNine}{17}

\newcommand{\posSZEightZNine}{10}

\newcommand{\meanSZEightZNine}{275}
\newcommand{\pSZEightZNine}{1.00}

\newcommand{\posOrigZEightZEight}{9}

\newcommand{\posMZEightZEight}{18}

\newcommand{\posSZEightZEight}{16}

\newcommand{\meanSZEightZEight}{180}
\newcommand{\pSZEightZEight}{0.01}

\newcommand{\sdWider}{8}
\newcommand{\nTasks}{9}
\newcommand{\pooledRho}{0.18}
\newcommand{\pooledP}{0.02}
\newcommand{\pooledN}{178}

\newcommand{\grokAll}{179}

\newcommand{\reproMult}{10}

\newcommand{\prevDSixteenGrok}{14}
\newcommand{\synthSplitZEightZNine}{0.52}
\newcommand{\synthMergedZEightZNine}{1.00}
\newcommand{\synthSplitZFourZFive}{0.58}
\newcommand{\synthMergedZFourZFive}{1.00}
\newcommand{\irrepHomErr}{3.2\times10^{-14}}
\newcommand{\irrepUniErr}{5.6\times10^{-16}}
\newcommand{\initAboveSFive}{2}
\newcommand{\initValsSFive}{0.74 and 0.56}
\newcommand{\initFirstCkptSFive}{0.36 and 0.11}
\newcommand{\initSyncSFive}{300 and 400}
\newcommand{\vOneNullStdMax}{3.3\times10^{-16}}
\newcommand{\dropBlockZeroMLP}{11.9}
\newcommand{\dropOthersMin}{88}
\newcommand{\eqRiseMinWins}{16}
\newcommand{\eqRiseMaxP}{0.012}
\newcommand{\eqLoDlog}{0.34}
\newcommand{\eqHiDlog}{0.59}
\newcommand{\eqLoNat}{0.15}
\newcommand{\eqHiNat}{0.40}
\newcommand{\riseDlogNinetySeven}{0.26}
\newcommand{\riseNatNinetySeven}{0.45}
\newcommand{\eqRiseDeltaLo}{0.10}
\newcommand{\eqRiseDeltaHi}{0.35}
\newcommand{\maxUntrainedNull}{0.40}
\newcommand{\untrainedOverThresh}{3}
\newcommand{\untrainedGroupNets}{1{,}200}

\newcommand{\winVsDsDSixteen}{3}
\newcommand{\lossVsDsDSixteen}{13}

\newcommand{\mRdZFiveZFive}{19}

\newcommand{\idealRdAbelianLo}{0.48}
\newcommand{\idealRdAbelianHi}{0.71}

\newcommand{\idealMultNat}{0.35}
\newcommand{\idealMultRd}{0.40}
\newcommand{\idealMultWrongMax}{0.57}

\newcommand{\sWinIrrDsSFive}{19}

\newcommand{\sPIrrDsSFive}{3.8\times10^{-6}}

\newcommand{\sWinDsRdSFive}{10}
\newcommand{\sLossDsRdSFive}{8}

\newcommand{\idealCycDsSFive}{0.07--0.29}
\newcommand{\idealCycRdSFive}{0.03--0.06}

\newcommand{\sWinDsRdDSixteen}{19}
\newcommand{\sLossDsRdDSixteen}{0}

\newcommand{\idealCycDsDSixteen}{0.61--0.67}
\newcommand{\idealCycRdDSixteen}{0.21--0.27}

\newcommand{\sigIrrRdZFiveZFive}{7}

\newcommand{\sigCells}{10}
\newcommand{\sFiveIrrDsFromSingle}{0.26}
\newcommand{\sFiveIrrDsFromHeld}{0.15}
\newcommand{\idealCycRdAbLo}{0.38}
\newcommand{\idealCycRdAbHi}{0.52}
\newcommand{\oneDShareDSixteen}{0.49}
\newcommand{\oneDShareSFive}{0.10}
\newcommand{\oneDUntrainedMeanDSixteen}{0.15}
\newcommand{\oneDUntrainedMaxDSixteen}{0.34}
\newcommand{\oneDAboveMaxDSixteen}{12}
\newcommand{\oneDNDSixteen}{19}
\newcommand{\effRankSyncSFive}{9.3}
\newcommand{\effRankSyncDSixteen}{3.3}
\newcommand{\effRankLateMax}{1.7}
\newcommand{\lateRandLo}{0.47}
\newcommand{\lateRandHi}{0.63}
\newcommand{\nullRdLo}{0.04}
\newcommand{\nullRdHi}{0.10}
\newcommand{\nullIrrLo}{0.24}
\newcommand{\nullIrrHi}{0.40}
\newcommand{\sPIrrDsOthersMin}{0.09}
\newcommand{\dSixteenAllMinP}{0.08}

\newcommand{\fsd}{\mathrm{FSD}_{\mathrm{gen}}}
\newcommand{\Z}{\mathbb{Z}}
\newtheorem{definition}{Definition}

\title{Learned Structure Shows Up Earlier in the Task's Own Basis:\\
       A Controlled Look at a Synchronisation Precursor in Grokking}
\author{Achyuthan Sivasankar\\
        New York University\\
        \texttt{as21154@nyu.edu}}
\date{}

\begin{document}
\maketitle

\begin{abstract}
Analyses of grokking usually read a network's hidden activations in one fixed basis, most often
the Fourier basis of the integers modulo $p$. We ask how much the answer to ``has the network
organised yet?''\ depends on that choice. We use $\fsd$, which averages each hidden unit's
activation over all inputs with the same answer and counts how many units share the same
strongest irreducible representation of the task's group. For modular multiplication we read one
and the same answer-averaged profile in three orders: discrete-logarithm order, in which the
characters of the cyclic multiplicative group become ordinary Fourier modes; the natural integer
order; and random orders. Across \multNets{} networks trained at $p\in\{59,97,113\}$, the
discrete-log reading reached a fixed threshold before both other readings in \bothMTotal{} of
\multNets{} networks, and in \bothSTotal{} of \multNets{} when the threshold had to hold for 500
steps. A fixed threshold favours the discrete-log reading, because untrained networks already score
higher in that order; when each reading must instead rise by the same amount above its own
untrained level, the discrete-log reading is still first in at least \eqRiseMinWins{} of 20
networks at every prime. At $p=59$ and $p=97$ this happened during the memorisation plateau, before the network
generalised, in \plateauSmallTotal{} of \firedSmallTotal{} networks; at $p=113$ it did so in
\plateauOneThirteen{} of 20. We also report what does not hold. Memorisation of the training set
preceded grokking in all \grokAll{} runs that grokked, and on average by a wider margin than any
synchronisation signal, so a lead over grokking is not by itself evidence that a circuit has been detected.
Synchronisation time is only weakly related to grokking time (pooled Spearman
$\rho=\pooledRho$). On a family of six finite groups, read with the same measure in every basis,
the reading in irreducible representations reached the threshold before random element orders on
all six. With equal-rise thresholds this held in every case on four groups, in
\sigIrrRdZFiveZFive{} of \sigCells{} cases on $\Z_5\times\Z_5$ and in none on $D_{16}$, whose
networks put about half of their profile power into one-dimensional representations that the
measure leaves out. Finally, ranking complex characters one at a time splits each unit's vote across
conjugate pairs and roughly halves the measure on abelian groups; we correct for this. This
version replaces an earlier one whose causal and predictive claims we withdraw.
\end{abstract}

\section{Introduction}

A small transformer trained on an algebraic task such as modular arithmetic often fits its
training set within a few hundred steps and reaches high test accuracy only thousands of steps
later \citep{power2022grokking}. Mechanistic studies of this \emph{grokking} have described what
the network ends up computing. For modular addition the generalising solution uses Fourier
features of $\Z_p$ \citep{nanda2023progress}. For composition in other finite groups it uses the
group's irreducible representations \citep{chughtai2023toy,stander2024grokking}, and different
runs on the same task can find different algorithms \citep{zhong2023clock}. Progress measures
built from these descriptions change long before test accuracy does \citep{nanda2023progress},
and hidden progress of a similar kind has been documented for parity learning
\citep{barak2022hidden}.

All of these analyses read the activations in some basis. For modular addition the Fourier basis
of $\Z_p$ is the obvious choice, and it is the basis the network uses. For other tasks the right
basis may be less obvious, and a measure read in a basis the network does not use can stay flat
after the network has organised. The basis therefore becomes part of any claim about when a
circuit forms: the claim depends on the network, and also on the coordinates we chose to look in.

Here we hold the network and the measure fixed and change only the basis. The measure, $\fsd$,
averages each hidden unit's activation over all inputs that share an answer, takes the Fourier
transform of that profile over the task's group, and counts how many units have the same strongest
irreducible representation. Modular multiplication gives a clean test. The multiplicative group
$\Z_p^\times$ is cyclic, so if the answers are listed in discrete-logarithm order its characters
become ordinary Fourier modes. Listing the same answers in their natural order, or in a random
order, gives readings of exactly the same profile in which the group structure is scrambled.

\begin{figure}[t]
  \centering
  \includegraphics[width=\linewidth]{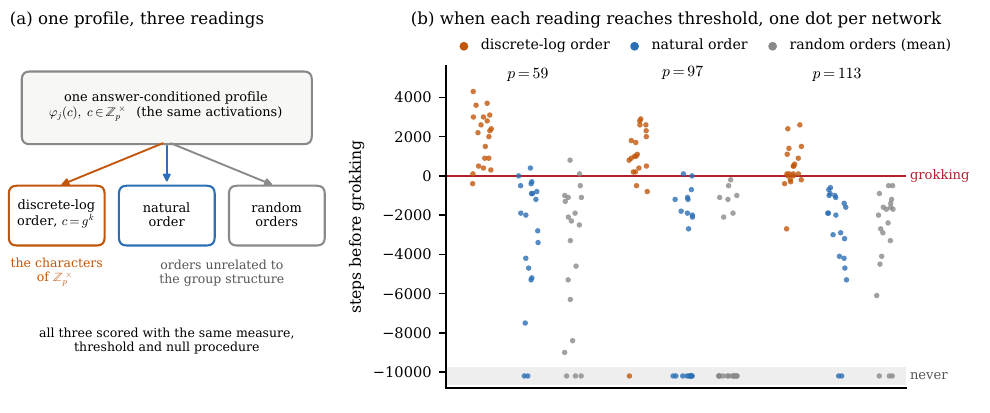}
  \caption{\textbf{One profile, three readings.} (a) For modular multiplication we average each
  hidden unit's activation over all inputs with the same answer $c\in\Z_p^\times$, then read this
  single profile in discrete-log order, in natural order and in random orders, scoring each
  reading with the same measure, threshold and null procedure. (b) For each of the
  \multNets{} networks, the step at which each reading reached the threshold and held it for 500
  steps, relative to the network's grokking step (positive means before grokking). The
  discrete-log reading came first in \bothSTotal{} of the \multNets{} networks.}
  \label{fig:overview}
\end{figure}

Our main finding is that the discrete-log reading reaches a fixed threshold before both other
readings in \bothMTotal{} of \multNets{} networks across three primes, and in \bothSTotal{} of
\multNets{} when the threshold must hold for 500 steps (Figure~\ref{fig:overview}). A fixed
threshold favours the discrete-log reading, whose untrained level is higher, so we also require an
equal rise above each reading's own untrained level; the discrete-log reading is then still first
in at least \eqRiseMinWins{} of 20 networks at every prime. At two of the
three primes this happens during the memorisation plateau, before the network generalises. Several
things do not hold. A lead over grokking is weak evidence on its own, because memorising the
training set precedes grokking on every run, and on average by more than any synchronisation signal.
Synchronisation time says little about when a particular run will grok. In a family of six
groups read with the same measure, the reading in irreducible representations comes before random
element orders on every group, but on $D_{16}$ only because of the fixed threshold.

This paper replaces an earlier version (arXiv:2606.12966v1), which described Fourier-basis
synchronisation on modular addition as a causal precursor of grokking and as the earliest
available predictor of circuit formation, and which read the late signal on modular
multiplication as specific to that operation. On re-examination we found that its significance
test was degenerate, that its multiplication comparison conditioned the activations on a different
variable, that its causal claim rested on an intervention that never manipulated synchronisation,
and that a lead over grokking does not separate synchronisation from memorisation. We re-ran every
experiment reported here with checkpoints kept, and fixed the synchronisation rules before the
sweeps; Section~\ref{sec:stats} says which analyses were chosen afterwards.
Appendix~\ref{app:changes} lists what we withdraw.

Our contributions are:
\begin{itemize}
  \item A basis-explicit synchronisation measure with two corrections: conjugate pairs of complex
        representations are merged before ranking, and the reference level of each reading is
        measured on untrained networks rather than on noise (Section~\ref{sec:method}).
  \item A controlled comparison on modular multiplication at three primes, in which only the order
        of the answers changes between readings, with a check that the result is not produced by
        the fixed threshold (Section~\ref{sec:mult}).
  \item The same comparison on six finite groups, with synthetic layers that show which element
        orders are wrong bases, and a partial diagnosis of the one group on which the effect does
        not survive (Section~\ref{sec:groups}).
  \item Two negative results: a lead over grokking does not distinguish this measure from
        memorisation, and synchronisation time is only weakly related to grokking time
        (Section~\ref{sec:timing}).
\end{itemize}

\section{Related work}

\paragraph{Grokking.} Delayed generalisation was reported by \citet{power2022grokking} and has
been studied as representation learning \citep{liu2022towards}, as an effect of weight norm and
initialisation scale \citep{liu2023omnigrok}, as competition between a memorising and a more
efficient generalising circuit \citep{varma2023grokking,merrill2023tale}, as a transition from
lazy to rich training dynamics \citep{kumar2024grokking,lyu2024dichotomy}, as a phase transition
\citep{rubin2024grokking}, and in relation to double descent \citep{davies2023unifying},
optimiser dynamics \citep{thilak2022slingshot} and label noise \citep{doshi2024grok}. Analyses of
modular arithmetic show how Fourier features arise \citep{gromov2023grokking} and why
margin maximisation favours them on algebraic tasks \citep{morwani2024feature}.

\paragraph{Progress measures and prediction.} \citet{nanda2023progress} built progress measures
from the reverse-engineered Fourier circuit, and \citet{barak2022hidden} documented hidden
progress in parity learning. \citet{notsawo2023predicting} predict grokking from early features of
the loss. Our measure is closest to the first kind: it needs to know which basis the circuit lives
in, though not the circuit itself. We find it is a poor predictor of grokking time, so it does not
compete with \citet{notsawo2023predicting} on that task.

\paragraph{Group structure in learned circuits.} \citet{chughtai2023toy} showed that networks
trained on group composition learn circuits built from the group's irreducible representations,
and \citet{stander2024grokking} described a coset-based algorithm for symmetric groups.
\citet{zhong2023clock} found that the algorithm can differ between runs. Group representations
also underlie equivariant architectures \citep{cohen2016group,kondor2018generalization}. The
harmonic analysis we use is standard \citep{serre1977linear,diaconis1988group}.

\paragraph{Bases in interpretability.} Whether a representation has a privileged basis is a
recurring question in interpretability \citep{olah2020zoom,elhage2022superposition}. Similarity
measures such as CKA are designed to be invariant to orthogonal changes of basis
\citep{kornblith2019similarity}. Concentration measures like ours are not, and this paper is about
how much that matters in practice.

\section{Setup and measure}
\label{sec:method}

\subsection{Tasks and training}

We train two-layer transformers \citep{vaswani2017attention} on binary operations
$(x,y)\mapsto x\star y$. The input is the token sequence $[x,y,=]$ and the model predicts the
answer at the last position. Each block applies layer normalisation, four-head self-attention,
layer normalisation and a GELU MLP of width 512, with residual connections, and
$d_\mathrm{model}=128$. We train with AdamW \citep{loshchilov2019decoupled} at learning rate
$10^{-3}$, weight decay $1.0$ and batch size 512, and save a checkpoint every 100 steps. Every
configuration is trained with 20 seeds (0--17, 42 and 123).

The tasks are composition in six finite groups and multiplication modulo a prime. The groups are
the symmetric group $S_5$ (order 120); the dihedral group $D_{16}$ of the regular 16-gon (order
32); two cyclic products, $\Z_4\times\Z_5\cong\Z_{20}$ and $\Z_8\times\Z_9\cong\Z_{72}$; and two
non-cyclic products, $\Z_5\times\Z_5$ and $\Z_8\times\Z_8$. Group tasks train on half of all pairs.
Multiplication uses $p\in\{59,97,113\}$ and trains on 30\% of pairs. Step budgets are listed in
Appendix~\ref{app:setup}; all 180 runs grokked except one $D_{16}$ run.

A run \emph{groks} at the first checkpoint with validation accuracy at least 0.95, and has
\emph{memorised} its training set at the first checkpoint with accuracy at least 0.99 on the full
training set. We measure the hidden layer of the first block's MLP (after the GELU) at the last
position.

\subsection{Answer-averaged profiles and the measure}

For a task whose answers lie in a group $G$, the \emph{profile} of hidden unit $j$ is its mean
activation over all inputs with a given answer,
\begin{equation*}
  \varphi_j(c) = \operatorname{mean}\bigl\{\mathrm{act}_j(x,y) : x\star y=c\bigr\},
  \qquad c\in G .
\end{equation*}
For each irreducible unitary representation $\rho$ of $G$, of dimension $d_\rho$, the Plancherel
power of the profile is
\begin{equation*}
  P_j(\rho) = \frac{d_\rho}{|G|}\,\Bigl\|\sum_{c\in G}\varphi_j(c)\,\rho(c)\Bigr\|_F^2 ,
\end{equation*}
and these powers sum to $\sum_c\varphi_j(c)^2$ \citep{serre1977linear}. On a cyclic group this is
the squared magnitude of the discrete Fourier transform.

\begin{definition}[$\fsd$]
Centre each profile over $c$ and divide each power by $d_\rho^2$, which is the expected power of
a random profile in $\rho$, so that every representation has the same expected normalised power.
If $G$ has
representations of dimension greater than one, rank only those; otherwise rank all non-trivial
ones. If $\rho$ is complex, so that its conjugate $\bar\rho$ is a different representation, replace
the pair by a single entry with power $P_j(\rho)+P_j(\bar\rho)$. Let $R$ be the resulting set of
entries, $r_j$ the entry with the largest power for unit $j$, and $\mathrm{par}(\rho)$ the
fraction of the $n$ units with $r_j=\rho$. Then
\begin{equation*}
  \fsd = \frac{\max_{\rho}\mathrm{par}(\rho)-1/|R|}{1-1/|R|}.
\end{equation*}
\end{definition}

The merge of conjugate pairs is new in this version. For a real-valued profile
$P_j(\rho)=P_j(\bar\rho)$ exactly, but in floating point the two agree only to about $10^{-14}$,
so ranking them separately lets rounding error decide which of the two a unit votes for, and the
vote splits across the pair. On $S_5$ and $D_{16}$ every irreducible representation is real and
the merge changes nothing. On the abelian groups every character of order greater than two is
complex. Without the merge, a layer in which every unit uses the same frequency scores only
\synthSplitZEightZNine{} on $\Z_8\times\Z_9$ and \synthSplitZFourZFive{} on $\Z_4\times\Z_5$,
instead of \synthMergedZEightZNine{} (Appendix~\ref{app:conj}).

Two other choices in the definition have consequences later. Ranking only the representations of
dimension greater than one on a non-abelian group is inherited from the earlier version, whose aim
was to detect structure that a one-dimensional Fourier reading would miss. It makes the measure
blind to units organised around one-dimensional representations, of which $S_5$ has two and
$D_{16}$ four. Second, a merged pair has twice the expected power of a real character under a
random profile. On the abelian groups and for multiplication, the real characters of order two (at
most three of the 10 to 56 ranked entries) are therefore less likely to win than the others, and
the chance level $1/|R|$ is slightly too low. We did not correct for this; the untrained null
includes it.

\paragraph{What the measure can see.} Suppose unit $j$ computes a product of matrix coefficients
of one representation, $\mathrm{act}_j(x,y)=\rho(x)_{ab}\,\rho(y)_{kl}$, a form that appears in
group-composition circuits \citep{chughtai2023toy}. Averaging over the inputs with $xy=c$ and
applying Schur orthogonality gives
\begin{equation*}
  \varphi_j(c) = \frac{\delta_{bk}}{d_\rho}\,\rho(c)_{al}.
\end{equation*}
When $b=k$ the profile is a matrix coefficient of $\rho$, all of its power lies in $\rho$, and the
unit votes for the representation it uses. When $b\neq k$ the profile is identically zero and the
unit is invisible to the measure. A low $\fsd$ can therefore mean that nothing has formed, that it
is being read in the wrong basis, or that averaging over answers cancels it. A high $\fsd$ means
that many units agree on one representation.

\subsection{Readings of one profile}

For multiplication modulo $p$ the answers of interest form $\Z_p^\times=\{1,\dots,p-1\}$, which is
cyclic of order $p-1$. We compute one profile from all pairs with $x,y\neq0$ (the discrete
logarithm is undefined at zero) and read it as a function on $\Z_{p-1}$ through a bijection
$\sigma:\Z_{p-1}\to\Z_p^\times$. The \emph{discrete-log} reading uses $\sigma(k)=g^k$ for a
primitive root $g$. Because this $\sigma$ is a group isomorphism, $\fsd$ over the characters of
$\Z_{p-1}$ then equals $\fsd$ over the characters of $\Z_p^\times$. The \emph{natural} reading uses
$\sigma(k)=k+1$, and the \emph{random} reading averages $\fsd$ over 20 fixed random bijections.
All three use the same activations, the same averaging, the same measure and the same threshold;
only the order in which the answers are listed differs. A wrong order is not blind in principle.
In a synthetic layer whose units all carry one character of $\Z_{97}^\times$ with random phases,
$\fsd$ is 1.00 in discrete-log order, and over the characters it has a median of \idealMultNat{}
in natural order and \idealMultRd{} in random orders (at most \idealMultWrongMax{}): the scrambled
profiles still lie in one two-dimensional space, and part of that agreement is registered. Our
intermediate drafts compared the discrete-log reading with activations averaged over $x+y$, which
is not the answer to a multiplication task. We keep that comparison only to check reproduction.

\subsection{Nulls and synchronisation rules}

For every task and reading we computed two null distributions of $\fsd$: one from profiles of
independent Gaussian noise, and one from 200 untrained networks whose initialisation seeds were
disjoint from the training seeds. The untrained null is the relevant one, and it is several times
higher (Appendix~\ref{app:nulls}). For $S_5$ its 95th percentile is \nullUntrainedSFive{}, against
\nullGaussSFive{} for Gaussian profiles.

A run \emph{synchronises} at the first checkpoint where $\fsd\geq0.50$ and $\fsd$ exceeds the 95th
percentile of the untrained null for that reading. Under the \emph{held} rule it must also stay
above both for 500 steps (six consecutive checkpoints). In practice the second condition never
applied: the untrained 95th percentile was below 0.50 for every task and reading (the largest was
\maxUntrainedNull{}, for $S_5$), so synchronisation is decided by the threshold alone. The
untrained level still matters, because it differs between readings, and a fixed threshold
therefore asks for a different rise from each; Section~\ref{sec:mult} checks what that does to the
comparison. The threshold of 0.50 is the value the earlier version used for $S_5$. We fixed it, the
rules, the seeds and the step budgets before running the sweeps; the budgets for $p=59$ and
$p=113$ came from a two-seed pilot. To check that the pipeline is unchanged, we also recompute
everything under the rule of our intermediate drafts, which ranked conjugate characters separately
and used the Gaussian null (the ``earlier rule'').

For the group family we read the same profile in two further ways, as a function on the cyclic
group $\Z_{|G|}$ through a listing of the elements: in the order the dataset uses, and averaged
over 20 fixed random orders. Both are scored with $\fsd$ over the characters of $\Z_{|G|}$, with
their own untrained nulls and the same rules, so that, as for multiplication, only the basis
changes. These readings were computed after the sweep from its stored checkpoints, and the
irreducible-representation trajectories they recompute agree exactly with the sweep's. Our
intermediate drafts used a different control, a top-5 participation score held to a stricter
threshold; Appendix~\ref{app:top5} reports it and why it misled. To see what each reading can
register, we also score synthetic \emph{ideal layers}, in which every unit is a random combination
of the matrix coefficients of one representation; $\fsd$ in the irreducible representations is
1.00 on all of them.

\subsection{Statistics}
\label{sec:stats}

Comparisons with grokking count the runs in which a signal came first, with a two-sided exact sign
test over the runs where the signal occurred. Paired comparisons between readings count the runs
in which one reading synchronised strictly before the other, or the other never did, with a
two-sided sign test over the runs where they differed. For timing we compute Spearman correlations
within each task, and a pooled correlation of within-task ranks, with two-sided permutation tests
(10{,}000 permutations, following \citealp{freedman1983nonstochastic}). We had planned to control
these correlations for memorisation time, but every run of a given task memorised at the same
checkpoint, so that control has no effect and the reported values are plain Spearman correlations.

\paragraph{What was planned.} The synchronisation rules and the counts of runs in which a signal
precedes grokking were specified before the sweeps were run, and the timing analysis was written
down before it was run on these data. The paired comparison between readings was first computed on
the $p=97$ runs; the runs at $p=59$ and $p=113$ were trained afterwards, so for those two primes it
is a replication. The same-measure readings of the group family were added after the sweep, to
replace a control we found to be flawed (Appendix~\ref{app:top5}); their rules and nulls are the
ones fixed for the sweep. Several analyses were chosen after we had seen the results and should be
read as exploratory: the classification of each synchronisation as before memorisation, inside the
plateau or after grokking; the equal-rise checks in Sections~\ref{sec:mult} and~\ref{sec:groups};
the ideal layers; and the diagnostics for $D_{16}$. We do not correct for multiple comparisons.

\section{Results}

\subsection{The pipeline reproduces our earlier sweep}

The group-family and discrete-log experiments were first run for an intermediate draft that we did
not post (Appendix~\ref{app:changes}); of this family the first version contains only a six-seed
$S_5$ experiment. Under that draft's rule, the re-run reproduces its counts and mean leads exactly
for $S_5$, $\Z_4\times\Z_5$ and $\Z_8\times\Z_9$, and for multiplication at $p=97$ it reproduces
the grokking and synchronisation steps of all \reproMult{} seeds that had been run before.
$D_{16}$ differs by design: we raised its step budget from 12{,}000 to 20{,}000 steps, so
\nGrokDSixteen{} of its 20 runs grokked rather than \prevDSixteenGrok{}. This shows that the
pipeline and its seeds are unchanged. It is not an independent replication.

\subsection{Same profile, three readings}
\label{sec:mult}

\begin{table}[t]
  \centering
  \caption{Modular multiplication. Number of runs in which each reading synchronised before
  grokking, out of the runs in which it synchronised at all; the number of runs in which it never
  synchronised is in parentheses. Mean leads are in steps.}
  \label{tab:mult}
  \small
  \begin{tabular}{@{}llcccc@{}}
\toprule
 & & \multicolumn{2}{c}{single crossing} & \multicolumn{2}{c}{held 500 steps} \\
\cmidrule(lr){3-4}\cmidrule(lr){5-6}
$p$ & reading of the same profile & leads grokking & mean lead & leads grokking & mean lead \\
\midrule
59 & discrete-log order & 20/20 & 2{,}850 & 19/20 & 1{,}960 \\
 & natural order & 9/20 & $-$105 & 1/18 (2) & $-$2{,}311 \\
 & random orders & 5/20 & $-$645 & 2/17 (3) & $-$2{,}929 \\
\addlinespace
97 & discrete-log order & 20/20 & 1{,}745 & 17/19 (1) & 1{,}237 \\
 & natural order & 3/19 (1) & $-$789 & 1/11 (9) & $-$1{,}327 \\
 & random orders & 0/17 (3) & $-$1{,}029 & 0/7 (13) & $-$1{,}143 \\
\addlinespace
113 & discrete-log order & 15/20 & 890 & 13/20 & 405 \\
 & natural order & 0/19 (1) & $-$1{,}511 & 0/18 (2) & $-$2{,}306 \\
 & random orders & 0/19 (1) & $-$1{,}832 & 0/17 (3) & $-$2{,}300 \\
\bottomrule
\end{tabular}

\end{table}

\begin{table}[t]
  \centering
  \caption{Modular multiplication, paired comparison. Left: runs (of 20) in which the
  discrete-log reading synchronised before both the natural and the random reading. Right: where
  the held discrete-log synchronisation falls relative to memorisation and grokking.}
  \label{tab:paired}
  \small
  \begin{tabular}{@{}lcccccc@{}}
\toprule
 & \multicolumn{2}{c}{discrete-log order reaches threshold first} & \multicolumn{4}{c}{where the held discrete-log sync falls} \\
\cmidrule(lr){2-3}\cmidrule(lr){4-7}
$p$ & single crossing & held 500 steps & before memorised & in plateau & at/after grokking & never \\
\midrule
59 & 20/20 & 20/20 & 1 & 18 & 1 & 0 \\
97 & 20/20 & 19/20 & 0 & 17 & 2 & 1 \\
113 & 20/20 & 20/20 & 1 & 12 & 7 & 0 \\
\bottomrule
\end{tabular}

\end{table}

Table~\ref{tab:mult} counts the runs in which each reading synchronised before grokking. In
discrete-log order it did so in all 20 runs at $p=59$ and $p=97$ under the single-crossing rule,
and in \posDlogMOneThirteen{} at $p=113$. The natural and random orders rarely did: never at
$p=113$, and in \posNatMNinetySeven{} and \posPermMNinetySeven{} runs at $p=97$. At $p=59$ the
natural order synchronised before grokking in \posNatMFiftyNine{} runs under the single-crossing
rule, though in only \posNatSFiftyNine{} under the held rule. The wrong orders are not blind; they
see the structure later and less reliably.

The paired comparison asks the more direct question, which reading of the same profile reached the
threshold first, and it does not depend on where grokking falls. The discrete-log reading came
first against both the natural and the random orders in all 20 runs at every prime under the
single-crossing rule, and in \bothSFiftyNine{}, \bothSNinetySeven{} and \bothSOneThirteen{} of 20
runs under the held rule (Table~\ref{tab:paired}, Figure~\ref{fig:overview}b). Against either
alternative alone the two-sided sign test gives $p<10^{-4}$ at every prime.

\paragraph{The fixed threshold favours the discrete-log reading.} Untrained networks score higher
in discrete-log order than in natural order (95th percentiles \nullDlogNinetySeven{} and
\nullNatNinetySeven{} at $p=97$), so averaging over answers already produces some structure in the
right basis before training. A threshold of 0.50 therefore asks the discrete-log reading to rise by
\riseDlogNinetySeven{} at $p=97$ and the natural reading by \riseNatNinetySeven{}, and some of the
head start in Table~\ref{tab:paired} could come from that alone. To check, we repeated the paired
comparison with each reading's threshold set at its own untrained 95th percentile plus a common
rise $\Delta$, for $\Delta$ from \eqRiseDeltaLo{} to \eqRiseDeltaHi{}
(Appendix~\ref{app:equalrise}; at $p=97$, $\Delta=0.26$ puts the discrete-log threshold at 0.50).
The discrete-log reading still came first in at least \eqRiseMinWins{} of 20 runs at every prime,
for every $\Delta$, under both rules and against either alternative (two-sided sign test
$p\leq\eqRiseMaxP$ in every case). An equal additive rise is the most favourable of the obvious
choices for the wrong orders; scaling the rise by the headroom above the null would set their
thresholds higher. The order between the readings is therefore not produced by the fixed
threshold, although the size of the lead under a fixed threshold partly is. No untrained network
reached 0.50 in discrete-log order (the largest value was \nullDlogMaxFiftyNine{}, at $p=59$).

\begin{figure}[t]
  \centering
  \includegraphics[width=\linewidth]{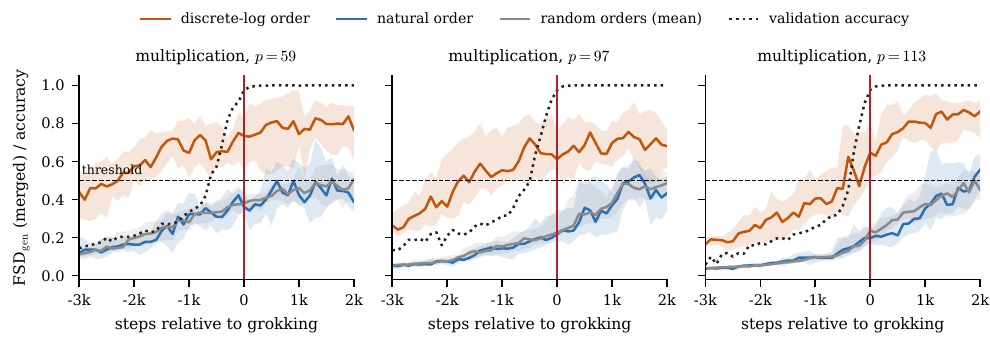}
  \caption{\textbf{Trajectories aligned at grokking.} Median $\fsd$ across the 20 runs at each
  prime for each reading, with interquartile bands, and median validation accuracy (dotted). The
  dashed line is the threshold of 0.50.}
  \label{fig:traj}
\end{figure}

\paragraph{When the structure becomes visible.} Figure~\ref{fig:traj} aligns every run at its
grokking step, and Table~\ref{tab:paired} places each held discrete-log synchronisation relative to
memorisation and grokking. At $p=59$ and $p=97$ it fell inside the plateau between them in
\plateauFiftyNine{} of \firedSFiftyNine{} and \plateauNinetySeven{} of \firedSNinetySeven{} runs:
the training set had been memorised (by steps \memStepFiftyNine{} and \memStepNinetySeven{}) and
the network had not yet generalised. At $p=113$ the runs grokked early, between steps
\grokMinOneThirteen{} and \grokMaxOneThirteen{}, and only \plateauOneThirteen{} of 20 held
synchronisations fell inside the plateau; in \afterGrokOneThirteen{} the structure became visible
in our measure only at or after grokking. On average the held discrete-log synchronisation led
grokking by \meanDlogSFiftyNine{}, \meanDlogSNinetySeven{} and \meanDlogSOneThirteen{} steps at the
three primes. The order between readings is the same at all three primes; whether the structure is
visible before generalisation depends on the prime.

\subsection{A family of groups}
\label{sec:groups}

\begin{table}[t]
  \centering
  \caption{Group composition. Runs in which $\fsd$ in the irreducible representations synchronised
  before grokking, out of the grokked runs in which it synchronised, under three rules. The
  ``earlier rule'' ranks conjugate characters separately and uses the Gaussian null; ``merged'' and
  ``held'' are this paper's rules. The last column counts the runs in which the training set was
  memorised before grokking.}
  \label{tab:groups}
  \small
  \setlength{\tabcolsep}{4pt}
  \begin{tabular}{@{}llccccc@{}}
\toprule
 & & & \multicolumn{3}{c}{FSD$_\mathrm{gen}$, irreducible representations} & \\
\cmidrule(lr){4-6}
group & type & grokked & earlier rule & merged & held & memorised \\
\midrule
$S_5$ & non-abelian & 20/20 & 20/20 & 20/20 & 20/20 & 20/20 \\
$D_{16}$ & non-abelian & 19/20 & 19/19 & 19/19 & 19/19 & 19/19 \\
$\mathbb{Z}_4{\times}\mathbb{Z}_5$ & cyclic, $\cong\mathbb{Z}_{20}$ & 20/20 & 17/20 & 20/20 & 19/20 & 20/20 \\
$\mathbb{Z}_5{\times}\mathbb{Z}_5$ & non-cyclic & 20/20 & 17/20 & 20/20 & 19/20 & 20/20 \\
$\mathbb{Z}_8{\times}\mathbb{Z}_9$ & cyclic, $\cong\mathbb{Z}_{72}$ & 20/20 & 6/20 & 17/20 & 10/20 & 20/20 \\
$\mathbb{Z}_8{\times}\mathbb{Z}_8$ & non-cyclic & 20/20 & 9/20 & 18/20 & 16/20 & 20/20 \\
\bottomrule
\end{tabular}

\end{table}

\begin{table}[t]
  \centering
  \caption{Group composition, read with $\fsd$ in three bases: the irreducible representations, and
  $\Z_{|G|}$ in the dataset's element order and in random orders (threshold 0.50 above each
  reading's own untrained null, held for 500 steps). Left: grokked runs in which each reading
  synchronised before grokking. Middle: runs in which the first-named reading reached the threshold
  first, against runs in which it was second, with the two-sided sign-test $p$. Right: the score of
  an ideal layer in the two $\Z_{|G|}$ readings, as a range over the ranked representations (real
  characters of order two score 1 in every order and are left out).}
  \label{tab:same}
  \small
  \setlength{\tabcolsep}{3pt}
  \begin{tabular}{@{}lcccccc@{}}
\toprule
 & leads grokking & \multicolumn{3}{c}{reaches the threshold first (wins--losses, $p$)} & \multicolumn{2}{c}{ideal layer} \\
\cmidrule(lr){3-5}\cmidrule(lr){6-7}
group & irr.\,/\,data.\,/\,rand. & irreps vs data. & irreps vs rand. & data. vs rand. & data. & rand. \\
\midrule
$S_5$ & 20\,/\,3\,/\,1 & 19--0 ($<$0.001) & 20--0 ($<$0.001) & 10--8 (0.81) & 0.07--0.29 & 0.03--0.06 \\
$D_{16}$ & 19\,/\,19\,/\,15 & 7--11 (0.48) & 18--1 ($<$0.001) & 19--0 ($<$0.001) & 0.61--0.67 & 0.21--0.27 \\
$\mathbb{Z}_4{\times}\mathbb{Z}_5$ & 19\,/\,20\,/\,7 & 7--5 (0.77) & 19--0 ($<$0.001) & 19--0 ($<$0.001) & 0.45--1.00 & 0.42--0.49 \\
$\mathbb{Z}_5{\times}\mathbb{Z}_5$ & 19\,/\,18\,/\,1 & 6--4 (0.75) & 19--1 ($<$0.001) & 20--0 ($<$0.001) & 1.00 & 0.42--0.52 \\
$\mathbb{Z}_8{\times}\mathbb{Z}_9$ & 10\,/\,8\,/\,1 & 10--3 (0.09) & 20--0 ($<$0.001) & 20--0 ($<$0.001) & 0.49--1.00 & 0.38--0.45 \\
$\mathbb{Z}_8{\times}\mathbb{Z}_8$ & 16\,/\,15\,/\,0 & 6--5 (1.00) & 19--1 ($<$0.001) & 20--0 ($<$0.001) & 0.48--1.00 & 0.38--0.46 \\
\bottomrule
\end{tabular}

\end{table}

\begin{figure}[t]
  \centering
  \includegraphics[width=\linewidth]{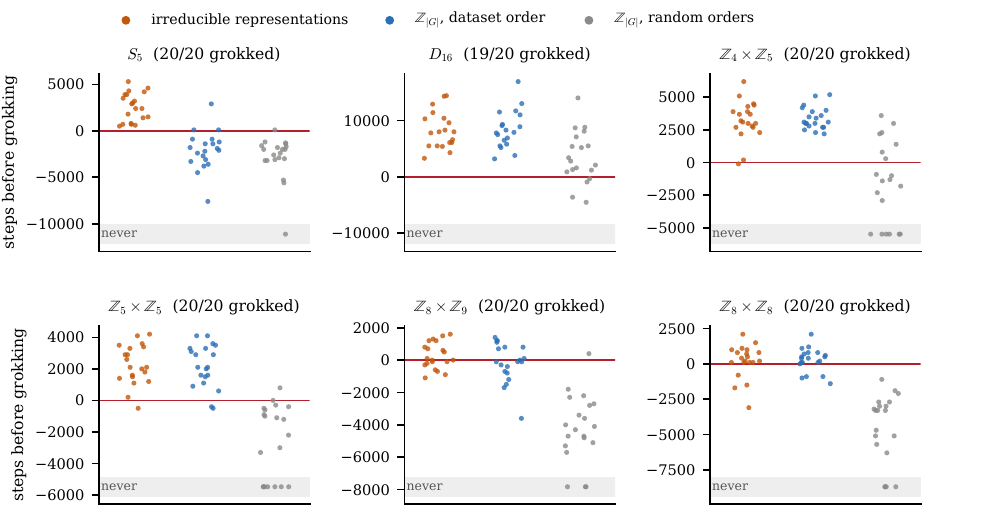}
  \caption{\textbf{Group composition.} Steps before grokking at which each reading synchronised
  under the held rule, one dot per grokked run: $\fsd$ in the irreducible representations, and
  $\fsd$ of the same profile read on $\Z_{|G|}$ in the dataset's element order and in random
  orders. Runs in which a reading never synchronised are shown in the grey band.}
  \label{fig:groups}
\end{figure}

Table~\ref{tab:groups} gives the counts for $\fsd$ in the irreducible representations.
Table~\ref{tab:same} and Figure~\ref{fig:groups} compare it with two readings of the same profile
as a function on $\Z_{|G|}$, scored with the same measure, so that only the basis changes.

\paragraph{Which orders are wrong bases?} The ideal layers answer this before any trained network
is considered (Table~\ref{tab:same}, right). In random orders an ideal layer scores
\idealCycRdSFive{} on $S_5$, \idealCycRdDSixteen{} on $D_{16}$ and between \idealCycRdAbLo{} and
\idealCycRdAbHi{} on the abelian groups, so a random order is a wrong basis on every group,
although on the abelian groups it still registers part of the agreement. The dataset's own order
is a wrong basis only on $S_5$ (\idealCycDsSFive{}). On $D_{16}$, which the dataset lists with its
16 rotations first, an ideal layer scores \idealCycDsDSixteen{}. On the abelian groups the reason
is the indexing. The dataset lists $(a,b)\in\Z_m\times\Z_n$ as $t=na+b$, and a character of the
second factor,
\begin{equation*}
  e^{2\pi\mathrm{i}\,lb/n} = e^{2\pi\mathrm{i}\,(lm)\,t/(mn)},
\end{equation*}
is an exact Fourier mode of the $mn$-point transform, so most ideal layers score 1.00 in dataset
order and the rest about 0.5.

\paragraph{Irreducible representations against random orders.} On every group, $\fsd$ in the
irreducible representations reached the held threshold before the random-order reading in all
or all but one of the runs (Table~\ref{tab:same}). This is the comparison that matches
multiplication, and it has the same weakness: a fixed threshold favours the reading whose
untrained level is higher, and here the gap is large. The untrained 95th percentiles of the
random-order readings lay between \nullRdLo{} and \nullRdHi{}, against \nullIrrLo{} to \nullIrrHi{}
for the irreducible representations. We therefore repeated the equal-rise check
(Appendix~\ref{app:groupequal}). The order survived every rise under both rules on $S_5$,
$\Z_4\times\Z_5$, $\Z_8\times\Z_9$ and $\Z_8\times\Z_8$, and in \sigIrrRdZFiveZFive{} of
\sigCells{} cases on $\Z_5\times\Z_5$. On $D_{16}$ it survived in none: there the two readings did
not differ significantly at any rise.

\paragraph{Irreducible representations against dataset order.} On $S_5$, where the dataset order is
a wrong basis, $\fsd$ in the irreducible representations came first in \sWinIrrDsSFive{} runs and
second in none ($p=\sPIrrDsSFive$). With an equal rise the difference is significant from
$\Delta=\sFiveIrrDsFromHeld$ under the held rule and from $\Delta=\sFiveIrrDsFromSingle$ under the
single-crossing rule, but not at the smallest rise. On the other five groups, where the dataset
order is in or near the right basis, the two readings did not differ ($p\geq\sPIrrDsOthersMin$).
The dataset order in turn came before the random orders on those five groups
(\sWinDsRdDSixteen{}--\sLossDsRdDSixteen{} on $D_{16}$ and at least 19--0 on each abelian group),
but not on $S_5$ (\sWinDsRdSFive{}--\sLossDsRdSFive{}). Across the family, then, a reading in or
near the group's own basis comes before a scrambled one, as it does for multiplication. The
exception is $D_{16}$ under equal rises.

\paragraph{$S_5$.} The timing differs from multiplication. In \beforeMemSFive{} of the 20 runs the
held synchronisation came before the training set was memorised at step \memStepSFive{}, so on
$S_5$ the units organise while the network is still fitting its training set. This is not left
over from initialisation. Only \initAboveSFive{} runs started above the threshold
(\initValsSFive{}); both had dropped to \initFirstCkptSFive{} by the first checkpoint and
synchronised later, at steps \initSyncSFive{}.

\paragraph{$D_{16}$.} The advantage of the irreducible representations over random orders on
$D_{16}$ comes from the fixed threshold. Part of the reason is visible in the profiles. At the held
synchronisation, on average \oneDShareDSixteen{} of the profile power of a $D_{16}$ network lay in
its one-dimensional representations, against \oneDUntrainedMeanDSixteen{} for untrained networks
(largest \oneDUntrainedMaxDSixteen{}); \oneDAboveMaxDSixteen{} of \oneDNDSixteen{} runs were above
every untrained network. $\fsd$ ranks only the two-dimensional representations of $D_{16}$, so it
reads about half of what these networks organise. Units that carry the same real one-dimensional
representation have the same profile up to scale, which every element order registers. Consistent
with this, the profile matrices of $D_{16}$ networks at synchronisation had an effective rank (the
participation ratio of their singular values) of \effRankSyncDSixteen{}, against
\effRankSyncSFive{} on $S_5$, where the one-dimensional share was \oneDShareSFive{}. Ranking the
one-dimensional representations as well moves the comparison towards the irreducible
representations at every rise, but not significantly (smallest $p=\dSixteenAllMinP$). On $D_{16}$,
then, we cannot conclude that the irreducible representations show the structure earlier.

\paragraph{Late in training the basis matters less.} By the end of training the profile matrices of
every group had a mean effective rank of at most \effRankLateMax{}: most units carried nearly the
same profile. Such a layer looks organised in any basis, and the random-order readings rose to
between \lateRandLo{} and \lateRandHi{} on average over the last checkpoints. The basis decides
when organisation becomes visible more than whether it eventually does, which is why we compare
readings by when they cross a threshold.

\paragraph{The larger abelian groups and the conjugate artefact.} Under the earlier rule only
\posOrigZEightZNine{} of 20 runs on $\Z_8\times\Z_9$ led grokking, and \posOrigZEightZEight{} of 20
on $\Z_8\times\Z_8$, which would make these groups look like failures. Most of this was the
conjugate artefact: with conjugate pairs merged the counts rise to \posMZEightZNine{} and
\posMZEightZEight{}. Under the held rule they fall back to \posSZEightZNine{} (sign test
$p=\pSZEightZNine$) and \posSZEightZEight{} ($p=\pSZEightZEight$), with mean leads of only
\meanSZEightZNine{} and \meanSZEightZEight{} steps. On these two groups the corrected signal is
present but weak and often brief.

\subsection{What synchronisation does not tell us}
\label{sec:timing}

\begin{figure}[!htb]
  \centering
  \includegraphics[width=\linewidth]{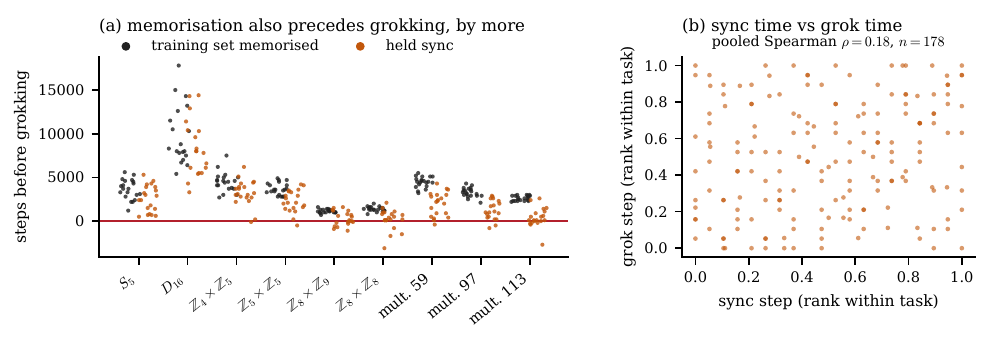}
  \caption{\textbf{What a lead does and does not show.} (a) Steps before grokking at which the
  training set was memorised and at which the held synchronisation occurred, one dot per run.
  Memorisation leads on every run. (b) Synchronisation step against grokking step, both ranked
  within each task, pooled over the \nTasks{} tasks.}
  \label{fig:timing}
\end{figure}

\begin{table}[!htb]
  \centering
  \caption{Synchronisation time and grokking time, held rule. $\rho$ is the Spearman correlation
  across seeds within a task, with a two-sided permutation $p$. Because every run of a task
  memorised at the same checkpoint, sd(grok$-$memorised) equals the spread of the grokking step.
  The last column places the held synchronisation before memorisation, inside the plateau, or at
  or after grokking.}
  \label{tab:timing}
  \small
  \setlength{\tabcolsep}{4pt}
  \begin{tabular}{@{}lcccccc@{}}
\toprule
task & $n$ & $\rho$ & $p$ & sd(grok$-$sync) & sd(grok$-$mem.) & before / in / after \\
\midrule
$S_5$ & 20 & $+0.53$ & $0.02$ & 1{,}518 & 1{,}142 & 10 / 10 / 0 \\
$D_{16}$ & 19 & $+0.17$ & $0.49$ & 3{,}267 & 3{,}377 & 0 / 19 / 0 \\
$\mathbb{Z}_4{\times}\mathbb{Z}_5$ & 20 & $+0.26$ & $0.27$ & 1{,}466 & 1{,}107 & 0 / 19 / 1 \\
$\mathbb{Z}_5{\times}\mathbb{Z}_5$ & 20 & $+0.11$ & $0.65$ & 1{,}261 & 684 & 0 / 19 / 1 \\
$\mathbb{Z}_8{\times}\mathbb{Z}_9$ & 20 & $-0.16$ & $0.50$ & 816 & 208 & 4 / 6 / 10 \\
$\mathbb{Z}_8{\times}\mathbb{Z}_8$ & 20 & $+0.15$ & $0.53$ & 1{,}190 & 271 & 1 / 15 / 4 \\
mult., $p=59$ & 20 & $+0.32$ & $0.16$ & 1{,}356 & 613 & 1 / 18 / 1 \\
mult., $p=97$ & 19 & $-0.08$ & $0.73$ & 1{,}110 & 549 & 0 / 17 / 2 \\
mult., $p=113$ & 20 & $+0.29$ & $0.22$ & 1{,}124 & 289 & 1 / 12 / 7 \\
\midrule
pooled & 178 & $+0.18$ & $0.02$ & & & \\
\bottomrule
\end{tabular}

\end{table}

\paragraph{A lead over grokking is weak evidence.} The training set was memorised before grokking
in all \grokAll{} runs that grokked, and in every task memorisation led grokking by a wider average
margin than any synchronisation rule (Figure~\ref{fig:timing}a). Any quantity that changes while
the network fits its training set will lead grokking. The earlier version treated a positive lead
as evidence that a circuit had been detected; on its own it is not. The comparisons between
readings in Sections~\ref{sec:mult} and~\ref{sec:groups} do not share this problem, because
memorisation is the same for every reading of the same activations.

\paragraph{Synchronisation time barely predicts grokking time.} Within a task, a run's held
synchronisation step was only weakly related to its grokking step (Table~\ref{tab:timing},
Figure~\ref{fig:timing}b). Pooled over the \nTasks{} tasks, the Spearman correlation of within-task
ranks was $\rho=\pooledRho$ ($n=\pooledN$, permutation $p=\pooledP$). The largest single-task
correlation was on $S_5$ ($\rho=\rhoSFive$, $p=\pRhoSFive$), which does not survive a correction for
\nTasks{} tasks; on multiplication at $p=97$ it was $\rho=\rhoMultNinetySeven$. Knowing the
synchronisation step did not narrow down the grokking step either: in \sdWider{} of the \nTasks{}
tasks the gap between synchronisation and grokking varied more across seeds than the grokking step
itself. Synchronisation tells us that structure has formed and in which basis it can be seen. It
does not tell us when a given run will generalise. The earlier version's weight-decay experiments
suggest one reason, which we have not re-examined here: the delay after the structure forms
depended on how quickly weight decay removed the memorising solution, and that process is not
visible in a measure of agreement between units.

\section{Discussion}

\paragraph{Reading circuits.} For multiplication, where we could change nothing but the order of
the answers, reading the activations in the basis of the task's own group made the learned
structure visible earlier in every network but one, and the order between the readings survived
replacing the fixed threshold by an equal rise above each reading's untrained level. Readings in
other orders were not blind, but they lagged, often until after the network had generalised. A flat or late signal from a
basis-dependent measure therefore says something about the basis as well as about the network, and
when the task's symmetry group is known it is a sensible default.

The group family adds three cautions about controls. First, a wrong-basis control is only as wrong
as its element order: in the order used to build our datasets, a cyclic reading of $D_{16}$ or of
an abelian product group is in or near the right basis. Second, a comparison between bases should
use one measure and one threshold. Our intermediate drafts did not, and that made the group family
look both cleaner and stranger than it is (Appendix~\ref{app:top5}). With one measure it agrees
with multiplication on every group except $D_{16}$, though less robustly on $\Z_5\times\Z_5$.
Third, a measure that ranks only some representations reads only part of the network: $D_{16}$
networks put about half of their profile power into representations that $\fsd$ leaves out.

\paragraph{Averaging over answers creates some structure.} Untrained networks score well above
Gaussian noise on our measure, and higher in the right basis than in the wrong one. When a unit's
activation is close to a product of functions of the two inputs, averaging over the inputs with a
given answer computes a convolution on the group, and convolution on a group is block-diagonal in
its Fourier basis (diagonal when the group is abelian); this may be why the concentration appears
before training, though we have not tested it. In either case, nulls for measures of this kind
should be computed from untrained networks rather than from noise, and because the untrained level
depends on the basis, a threshold shared between bases is not neutral between them.

\paragraph{Limitations.} Our models are small, with two layers and $d_\mathrm{model}=128$, and our
groups and primes are small. We measure one layer, the first block's MLP. In a zero-ablation
experiment on a single modular-addition run from the earlier work, removing this layer reduced
accuracy by \dropBlockZeroMLP{} percentage points, against at least \dropOthersMin{} for each other
component, so the structure we measure may accompany the main computation rather than carry it;
we did not repeat the ablation on the tasks studied here. The threshold of 0.50 was taken from
earlier work and not tuned, and apart from the equal-rise checks we did not study other
thresholds. Several analyses were chosen after the results were known
(Section~\ref{sec:stats}), including the
same-measure readings of the group family, which were computed from the stored checkpoints after
the sweep. Checkpoints
every 100 steps limit timing resolution, and they are why memorisation could not be used as a
covariate. We did not intervene on synchronisation, so we make no causal claim. The result on
$D_{16}$ is only partly explained, and the weaker timing at $p=113$ is unexplained. $\fsd$ does
not rank the one-dimensional representations of a non-abelian group, which on $D_{16}$ carry about
half of the profile power.

\paragraph{Predicting grokking.} A measure that predicts when a run will grok would need to track
the decay of the memorising solution as well as the formation of the generalising one, and should
be judged by its error on held-out runs. \citet{notsawo2023predicting} take a different approach to
this problem. Our results suggest that agreement between units in the first layer is not enough
on its own.

\section{Conclusion}

Holding the network and the measure fixed and changing only the basis, we found that learned
structure in networks trained on modular multiplication becomes visible earlier when activations
are read in the basis of the task's group. This held in \bothSTotal{} of \multNets{} networks
across three primes and did not depend on the fixed threshold, and at two of the three primes the
structure appeared during the memorisation plateau. Read with one measure, a family of six finite groups showed the same ordering on every
group except $D_{16}$, where it did not survive an equal-rise threshold and about half of the
profile power lies in representations our measure leaves out.
We also found that a lead over grokking is weak evidence for any progress measure, that
our measure does not predict when a network will grok, and that a basis-dependent control can be
partly in the right basis because of how the elements happen to be ordered. We hope these checks
are useful to others building measures of this kind.

\paragraph{Reproducibility.} Every number, table and figure in this paper is generated from the
per-run result files by a single script, which also verifies that the result files are complete.
Scripts and per-run results are available from the author.

\bibliographystyle{plainnat}
\bibliography{main}

\clearpage
\appendix

\section{Changes from the first version}
\label{app:changes}

The first version (arXiv:2606.12966v1) studied Fourier-basis synchronisation on modular addition
and a basis-faithful version on $S_5$. We withdraw or revise the following.

\begin{itemize}
  \item \emph{Causal claim.} The first version described synchronisation as a causal precursor of
        grokking and the circuit as computation-complete thousands of steps before grokking. Its
        evidence was an intervention that changed weight decay after synchronisation and never
        manipulated synchronisation. We withdraw the causal claim.
  \item \emph{Earliest predictor.} The first version called its measure the earliest available
        predictor of Fourier circuit formation. Memorisation of the training set precedes it, and
        it predicts grokking time poorly (Section~\ref{sec:timing}).
  \item \emph{Significance test.} The first version described its measure as permutation-tested.
        Its label-shuffle permutation null was degenerate: across the stored checkpoints of the
        main run, its standard deviation was at most $\vOneNullStdMax$. We replace it with null
        distributions from untrained networks.
  \item \emph{Multiplication.} The first version read the late signal on modular multiplication as
        evidence that the precursor was specific to addition. That comparison averaged activations
        over $x+y$, which is not the answer to a multiplication task. Section~\ref{sec:mult}
        replaces it with readings of one answer-averaged profile.
  \item \emph{$S_5$.} The first version reported a positive lead on six $S_5$ seeds while the
        Fourier reading did not lead; on one of those seeds the Fourier reading also led. We re-ran
        $S_5$ with 20 seeds and the corrected null (Section~\ref{sec:groups}).
  \item \emph{Not carried over.} The first version's analyses of modular addition and subtraction,
        its comparison with a restricted-logit loss, its weight-decay timing law and schedule, its
        symbolic extraction and its architecture ablations are not part of this version, and we
        have not re-examined them with the procedures used here. A reader of the first version
        pointed out numerical inconsistencies in some of them, such as synchronisation steps that
        differ between tables for the same run. We do not rely on any of these results.
\end{itemize}

Intermediate drafts, which were not posted, extended the work to a group family and to the
discrete-log reading. Their errors are corrected in this version. They include the conjugate
artefact, an incorrect claim that $\fsd$ locks onto multi-dimensional representations of
$\Z_4\times\Z_5$ (which has none), and a naive control that was held to a stricter threshold
than the measure it was compared with (Section~\ref{sec:groups}).

\section{Training and evaluation details}
\label{app:setup}

\begin{table}[h]
  \centering
  \small
  \begin{tabular}{@{}ll@{}}
    \toprule
    architecture & 2 blocks, $d_\mathrm{model}=128$, 4 heads, MLP width 512 (GELU), pre-layer-norm \\
    optimiser & AdamW, learning rate $10^{-3}$, weight decay $1.0$, batch 512 \\
    seeds & 0--17, 42, 123 \\
    training fraction & 0.5 (groups), 0.3 (multiplication) \\
    step budget & $S_5$ 12{,}000; $D_{16}$ 20{,}000; abelian groups 8{,}000; \\
                & multiplication $p=97$ 6{,}000, $p=113$ 10{,}000, $p=59$ 20{,}000 \\
    checkpoints & every 100 steps \\
    grokked & first checkpoint with validation accuracy $\geq0.95$ \\
    memorised & first checkpoint with full-training-set accuracy $\geq0.99$ \\
    measured layer & block-0 MLP, after the GELU, last position \\
    $\fsd$ & $K=1$, threshold 0.50, conjugate pairs merged \\
    untrained null & 200 initialisations per task and reading, seeds disjoint from training \\
    held rule & above threshold and null for 500 steps (six checkpoints) \\
    $\Z_{|G|}$ readings (groups) & $\fsd$ over the characters of $\Z_{|G|}$; dataset order and the \\
                & mean over 20 random orders; same threshold, nulls and rules \\
    top-5 control (App.~\ref{app:top5}) & $|G|$-point DFT, top-5 participation, threshold 0.50 held 500 steps \\
    ideal layer & 512 units, random combinations of one representation's matrix coefficients \\
    \bottomrule
  \end{tabular}
\end{table}

We construct the irreducible representations exactly: Young's orthogonal form for $S_5$, the
standard one- and two-dimensional representations for $D_{16}$, and products of characters for the
abelian groups. For every group we check that $\sum_\rho d_\rho^2=|G|$, that each representation is
unitary (largest error $\irrepUniErr$), and that $\rho(g)\rho(h)=\rho(gh)$ for all pairs (largest
error $\irrepHomErr$).

\section{The conjugate-pair artefact}
\label{app:conj}

To isolate the artefact we built a synthetic layer of 512 units on $\Z_8\times\Z_9$ in which every
unit computes the same real frequency, $\cos\bigl(2\pi(3a/8+4b/9)+\theta_j\bigr)$, with random
phases $\theta_j$ and amplitudes. Every unit uses one frequency, so the ideal value of the measure
is 1. Ranking the two complex characters of that frequency separately gives
\synthSplitZEightZNine{}, because rounding error splits the units between them; merging them gives
\synthMergedZEightZNine{}. The same construction on $\Z_4\times\Z_5$ gives \synthSplitZFourZFive{}
and \synthMergedZFourZFive{}. With a threshold of 0.50, the unmerged measure places a perfectly
organised abelian layer at the threshold. On $S_5$ and $D_{16}$ every representation is real, the
merge has no effect, and the counts under the merged and earlier rules agree
(Table~\ref{tab:groups}).

\section{Null distributions}
\label{app:nulls}

\begin{table}[h]
  \centering
  \caption{Null distributions of $\fsd$ (conjugate pairs merged). The Gaussian null uses profiles
  of independent standard normal noise; the untrained null uses 200 untrained networks.}
  \small
  \begin{tabular}{@{}llcccc@{}}
\toprule
task & reading & Gaussian $p_{95}$ & untrained $p_{95}$ & untrained mean & untrained max \\
\midrule
$S_5$ & irreps & 0.080 & 0.397 & 0.204 & 0.490 \\
$D_{16}$ & irreps & 0.050 & 0.362 & 0.189 & 0.635 \\
$\mathbb{Z}_4{\times}\mathbb{Z}_5$ & irreps & 0.047 & 0.343 & 0.193 & 0.514 \\
$\mathbb{Z}_5{\times}\mathbb{Z}_5$ & irreps & 0.037 & 0.308 & 0.182 & 0.497 \\
$\mathbb{Z}_8{\times}\mathbb{Z}_9$ & irreps & 0.024 & 0.265 & 0.144 & 0.341 \\
$\mathbb{Z}_8{\times}\mathbb{Z}_8$ & irreps & 0.027 & 0.239 & 0.146 & 0.386 \\
\addlinespace
mult., $p=59$ & discrete-log & 0.027 & 0.258 & 0.145 & 0.389 \\
 & natural & & 0.068 & & \\
 & random & & 0.060 & & \\
mult., $p=97$ & discrete-log & 0.023 & 0.238 & 0.129 & 0.370 \\
 & natural & & 0.051 & & \\
 & random & & 0.047 & & \\
mult., $p=113$ & discrete-log & 0.022 & 0.195 & 0.116 & 0.346 \\
 & natural & & 0.049 & & \\
 & random & & 0.040 & & \\
\bottomrule
\end{tabular}

\end{table}

The 95th percentile of the untrained null was below the threshold of 0.50 for every task and
reading, so the null condition in the synchronisation rule never changed a result. Individual
untrained networks did exceed 0.50 in the group family (\untrainedOverThresh{} of
\untrainedGroupNets{}); none did in discrete-log order for multiplication. For the random reading
of multiplication the untrained null was computed from the first four of the 20 random orders,
which makes its 95th percentile slightly higher than it would be with all 20.

\section{Equal-rise thresholds for multiplication}
\label{app:equalrise}

\begin{table}[h]
  \centering
  \caption{Paired comparison for multiplication when each reading's threshold is its own untrained
  95th percentile plus a common rise $\Delta$. Each cell gives the runs in which the discrete-log
  reading reached its threshold before the natural order, against the runs in which it was later,
  and then the same against the random orders. Ties are left out, so the two numbers need not sum
  to 20.}
  \label{tab:equalrise}
  \small
  \setlength{\tabcolsep}{5pt}
  \begin{tabular}{@{}llccccc@{}}
\toprule
 & & \multicolumn{5}{c}{required rise $\Delta$ above each reading's own untrained $p_{95}$} \\
\cmidrule(lr){3-7}
$p$ & rule & 0.10 & 0.15 & 0.20 & 0.26 & 0.35 \\
\midrule
59 & single & 16--4, 18--0 & 20--0, 20--0 & 20--0, 20--0 & 19--0, 20--0 & 20--0, 20--0 \\
 & held & 20--0, 20--0 & 19--1, 19--1 & 18--2, 18--2 & 18--1, 16--2 & 16--3, 18--1 \\
97 & single & 19--0, 19--0 & 19--0, 19--0 & 19--0, 20--0 & 20--0, 20--0 & 18--2, 20--0 \\
 & held & 20--0, 20--0 & 20--0, 20--0 & 20--0, 20--0 & 19--1, 18--1 & 18--1, 17--1 \\
113 & single & 16--2, 18--0 & 16--0, 19--0 & 16--1, 19--0 & 17--1, 19--0 & 19--0, 20--0 \\
 & held & 17--1, 17--1 & 20--0, 19--1 & 19--0, 20--0 & 19--0, 20--0 & 20--0, 20--0 \\
\bottomrule
\end{tabular}

\end{table}

The thresholds this produces at $p=97$ run from \eqLoDlog{} to \eqHiDlog{} for the discrete-log
reading and from \eqLoNat{} to \eqHiNat{} for the natural order. The smaller rises matter most for
the wrong orders, because a layer organised on a single character scores only about
\idealMultNat{} to \idealMultRd{} in them (Section~\ref{sec:method}), so their larger thresholds
are close to the most they can register.

\clearpage
\section{Equal-rise thresholds for the group family}
\label{app:groupequal}

Table~\ref{tab:groupequal} gives the equal-rise check of Section~\ref{sec:groups} for every group,
rule and rise.

\begin{table}[h]
  \centering
  \caption{Group family, with each reading's threshold at its own untrained 95th percentile plus a
  common rise $\Delta$. Each cell gives the runs in which $\fsd$ in the irreducible representations
  reached its threshold before the dataset-order reading, against the runs in which it was later,
  and then the same against the random orders. $^*$: two-sided sign test $p<0.05$ in favour of the
  irreducible representations.}
  \label{tab:groupequal}
  \small
  \setlength{\tabcolsep}{4pt}
  \begin{tabular}{@{}llccccc@{}}
\toprule
 & & \multicolumn{5}{c}{required rise $\Delta$ above each reading's own untrained $p_{95}$} \\
\cmidrule(lr){3-7}
group & rule & 0.10 & 0.15 & 0.20 & 0.26 & 0.35 \\
\midrule
$S_5$ & single & 11--9, 17--2$^*$ & 12--7, 19--0$^*$ & 13--7, 19--1$^*$ & 16--4$^*$, 19--1$^*$ & 19--1$^*$, 19--1$^*$ \\
 & held & 12--8, 17--2$^*$ & 17--3$^*$, 17--1$^*$ & 19--1$^*$, 19--1$^*$ & 18--1$^*$, 19--1$^*$ & 19--1$^*$, 19--1$^*$ \\
$D_{16}$ & single & 4--11, 7--11 & 5--13, 8--10 & 5--12, 8--9 & 6--12, 8--10 & 5--12, 11--7 \\
 & held & 5--11, 7--11 & 5--13, 10--9 & 7--11, 10--9 & 7--10, 11--8 & 10--8, 11--8 \\
$\mathbb{Z}_4{\times}\mathbb{Z}_5$ & single & 5--1, 12--2$^*$ & 4--3, 15--1$^*$ & 5--3, 16--1$^*$ & 7--5, 17--1$^*$ & 4--5, 17--1$^*$ \\
 & held & 4--2, 12--3$^*$ & 6--5, 15--3$^*$ & 6--7, 14--4$^*$ & 7--6, 14--3$^*$ & 7--4, 16--3$^*$ \\
$\mathbb{Z}_5{\times}\mathbb{Z}_5$ & single & 6--5, 13--4$^*$ & 4--6, 16--4$^*$ & 3--9, 16--4$^*$ & 3--10, 14--4$^*$ & 9--3, 17--1$^*$ \\
 & held & 5--7, 14--6 & 6--6, 16--4$^*$ & 4--6, 16--4$^*$ & 6--7, 12--6 & 6--8, 11--8 \\
$\mathbb{Z}_8{\times}\mathbb{Z}_9$ & single & 6--1, 19--0$^*$ & 7--4, 19--0$^*$ & 8--8, 20--0$^*$ & 8--5, 20--0$^*$ & 8--4, 18--0$^*$ \\
 & held & 6--4, 18--1$^*$ & 7--4, 20--0$^*$ & 7--5, 19--0$^*$ & 9--4, 20--0$^*$ & 8--6, 16--4$^*$ \\
$\mathbb{Z}_8{\times}\mathbb{Z}_8$ & single & 2--6, 19--0$^*$ & 3--4, 19--0$^*$ & 5--3, 20--0$^*$ & 6--5, 20--0$^*$ & 3--5, 20--0$^*$ \\
 & held & 3--6, 18--2$^*$ & 3--6, 20--0$^*$ & 5--6, 20--0$^*$ & 6--5, 18--1$^*$ & 3--6, 15--2$^*$ \\
\bottomrule
\end{tabular}

\end{table}

\section{The top-5 control of the intermediate drafts}
\label{app:top5}

\begin{table}[!htb]
  \centering
  \caption{Group composition, with the naive reading held to the threshold and persistence of
  $\fsd$ (0.50 for 500 steps). Left: grokked runs in which the naive reading came before grokking,
  in the dataset's element order and in random orders. Middle: runs in which $\fsd$ in the
  irreducible representations reached the threshold before the naive reading, against runs in
  which it was later, with the two-sided sign-test $p$. Right: the naive score of an ideal layer,
  as a range over the ranked representations (real characters of order two score 1 in every
  order and are left out).}
  \label{tab:matched}
  \small
  \setlength{\tabcolsep}{4pt}
  \begin{tabular}{@{}lcccccc@{}}
\toprule
 & \multicolumn{2}{c}{naive leads grokking} & \multicolumn{2}{c}{irreps first (wins--losses, $p$)} & \multicolumn{2}{c}{naive score, ideal layer} \\
\cmidrule(lr){2-3}\cmidrule(lr){4-5}\cmidrule(lr){6-7}
group & dataset & random & vs dataset & vs random & dataset & random \\
\midrule
$S_5$ & 6/20 & 1/20 & 19--1 ($<$0.001) & 19--0 ($<$0.001) & 0.17--0.50 & 0.07--0.11 \\
$D_{16}$ & 19/19 & 19/19 & 3--13 (0.02) & 11--7 (0.48) & 1.00 & 0.34--0.41 \\
$\mathbb{Z}_4{\times}\mathbb{Z}_5$ & 20/20 & 19/20 & 1--13 (0.002) & 14--3 (0.01) & 1.00 & 0.57--0.67 \\
$\mathbb{Z}_5{\times}\mathbb{Z}_5$ & 20/20 & 19/20 & 0--17 ($<$0.001) & 16--4 (0.01) & 1.00 & 0.57--0.71 \\
$\mathbb{Z}_8{\times}\mathbb{Z}_9$ & 20/20 & 2/20 & 0--18 ($<$0.001) & 20--0 ($<$0.001) & 1.00 & 0.48--0.61 \\
$\mathbb{Z}_8{\times}\mathbb{Z}_8$ & 20/20 & 2/20 & 0--18 ($<$0.001) & 18--1 ($<$0.001) & 1.00 & 0.50--0.65 \\
\bottomrule
\end{tabular}

\end{table}

Our intermediate drafts, and an earlier revision of Section~\ref{sec:groups}, compared $\fsd$ with
a different control: a $|G|$-point discrete Fourier transform of the profile, scored by the share
of units that have the most popular frequency among their five strongest, normalised by chance in
the same way as $\fsd$. The drafts held it to a threshold of 0.80 while $\fsd$ was held to 0.50.
Table~\ref{tab:matched} shows it at the matched threshold of 0.50 held for 500 steps. Two things
made it misleading. Its threshold was stricter than that of the measure it was compared with: at
0.80 the random-order reading led grokking in only \posZpPermZFiveZFive{} of
\firedZpPermZFiveZFive{} runs on $\Z_5\times\Z_5$, which the drafts read as random orders removing
the signal, while at 0.50 it led in \mRdZFiveZFive{} of 20. And top-5 participation is more
lenient than counting each unit's single strongest entry: in dataset order it scores an ideal layer
at 1.00 on $D_{16}$ and on every abelian group, and in random orders between \idealRdAbelianLo{}
and \idealRdAbelianHi{} on the abelian groups (Table~\ref{tab:matched}, right). With this control
the dataset-order reading appeared to come before $\fsd$ on $D_{16}$ (\lossVsDsDSixteen{} runs
against \winVsDsDSixteen{}) and on the abelian groups. With $\fsd$ in every basis
(Table~\ref{tab:same}) these differences disappear.

\end{document}